%% file: main.tex
\documentclass{article}
\usepackage[utf8]{inputenc}
\usepackage{booktabs}
\usepackage{geometry}
\usepackage{tikz}
\usepackage{multirow}
\usepackage{arydshln}
\usepackage{graphicx}
\usepackage{float}
\usepackage{todonotes}
\usepackage{amssymb}
\usepackage{amsmath}
\usepackage[numbers]{natbib}

\title{Fine-Tuning with Differential Privacy Necessitates\\ an Additional Hyperparameter Search}
\author{
Yannis Cattan$^\dagger$\footnote{Work partially completed while the author was at the Vector Institute.}~, Christopher A. Choquette-Choo$^+$, Nicolas Papernot$^+$, Abhradeep Thakurta$^+$\\
$^\dagger$Mines Paris, Université PSL, $^+$Google Research, Brain Team\\
cattan.yannis@gmail.com, $\{$cchoquette, papernot, athakurta$\}$@google.com
}
\date{}

\input{macros}

\begin{document}

\maketitle

\input{abstract}
\input{intro}
\input{background}

\input{method}
\input{experiments}
\input{discussion}

\bibliographystyle{unsrtnat}
\bibliography{biblio}

\end{document}

%% file: macros.tex
\newif\ifdraft
 \drafttrue

\definecolor{darkgreen}{rgb}{0,0.4,0.0}
\definecolor{darkblue}{rgb}{0,0.0,0.4}
\ifdraft
\newcommand{\cc}[1]{[{\color{darkgreen}{\textbf{CC:} \emph{#1}}}]}
\else
\newcommand{\cc}[1]{}
\fi

%% file: abstract.tex
\begin{abstract}
Models need to be trained with privacy-preserving learning algorithms to prevent leakage of possibly sensitive information contained in their training data. However, canonical algorithms like differentially private stochastic gradient descent (DP-SGD) do not benefit from model scale in the same way as non-private learning. This manifests itself in the form of unappealing tradeoffs between privacy and utility (accuracy) when using DP-SGD on complex tasks. To remediate this tension, a paradigm is emerging: fine-tuning with differential privacy from a model pretrained on public (i.e., non-sensitive) training data. 

In this work, we identify an oversight of existing approaches for differentially private fine tuning. They do not tailor the fine-tuning approach to the specifics of learning with privacy. Our main result is to show how carefully selecting the layers being fine-tuned in the pretrained neural network allows us to establish new state-of-the-art tradeoffs between privacy and accuracy. For instance, we achieve 77.9\% accuracy for $(\varepsilon, \delta)=(2, 10^{-5})$ on CIFAR-100 for a model pretrained on ImageNet.
Our work calls for additional hyperparameter search to configure the differentially private fine-tuning procedure itself.\footnote{The authors note that this manuscript is the preprint of an upcoming full version.} 
\end{abstract}

%% file: intro.tex
\section{Introduction}
Differentially private stochastic gradient descent (DP-SGD)~\cite{abadi2016deep} is the seminal algorithm for private learning~\cite{song2013stochastic,chaudhuri2011differentially}. However, its formulation also suffers from a type of `curse of dimensionality' that limits DP learning from benefiting from model scale (in the number of parameters $d$) in the same way as non-private learning. A likely explanation is because the norm of the noise required for DP guarantees in DP-SGD, grows as $\sqrt{d}$~\cite{bassily2014private,yu2021not}. Thus, though increasing $d$ can improve the model's representational power, it also increases the norm of the noise added. Indeed, increasing parameters in DP learning can even hurt model performance in practice~\cite{papernot2019making,shen2021towards,tramer2020differentially,yu2021not,kurakin2022toward}, unlike the performance increases observed in non-private learning~\cite{kaplan2020scaling,brown2020language,Zhai_2022_CVPR}.

When privacy is of concern, 
DP fine tuning 
has newly arisen as a method to overcoming this curse of dimensionality~\cite{li2021large,yu2021differentially,de2022unlocking,mehta2022large}. It works by leveraging large amounts of (generic) public data to train an initial non-private model; then, the model can be fine-tuned on the task-specific data. This paradigm initially emerged in the natural language processing (NLP) community, where large pretrained transformers were fine-tuned for downstream tasks~\cite{devlin2018bert}. Often, this achieves significant performance benefits compared with not leveraging the public data and has been used by the recent work of De et al.~\cite{de2022unlocking} to achieve state-of-the-art privacy-accuracy tradeoffs. For example, they achieve $84.8\%$ top-1 accuracy under $(0.5, 8\cdot10^{-7})$-DP %
on ImageNet by DP fine-tuning a model pretrained on the JFT dataset.

In our work, we ask whether fine tuning is always beneficial to DP learning. We find that prior work has so far overlooked the need to carefully attend to the specificity of private learners when designing the fine-tuning approach.
Our main result stems from a simple experiment: restricting the layers being fine-tuned with differential privacy. 
\textbf{In particular, we observe that simply fine-tuning both the first and last layers of a model---which amounts to a single line change in most libraries---consistently improves the final model performance, even with an implementation of DP-SGD already heavily tuned to achieve (previous) state-of-the-art performance.}

Because of this overlook in prior work, we are able to achieve new state-of-the-art performance without any other modifications than choosing to fine-tune the first and last layers of the pretrained model.
For instance, we apply this to a 28-10 ResNet on CIFAR-100 and find that this performs far superior to both of the prior DP finetuning methods: tuning the whole model or just the last layer. Our approach yields a model with 77.9\% performance at $(2,10^{-5})$-DP compared to 74.7\% (a $3.2$ percentage point improvement).

%% file: background.tex
\section{Differential Privacy}
\label{sec:background}

Before we delve into the specifics of private fine tuning, we provide an overview of differential privacy (DP) within the context of machine learning. 
DP is the gold standard for reasoning about privacy leakage through randomized queries of datasets. A query can be thought of as an outcome from a randomized mechanism $\mathcal{M}$ with domain $\mathcal{D}$ and range $\mathcal{R}$. $\mathcal{M}$ is said to satisfy $(\varepsilon, \delta)$-DP if for any subset $\mathcal{S} \subseteq \mathcal{R}$ and \emph{any} `adjacent' datasets $d, d' \in \mathcal{D}$ satisfying $\|d - d'\|_{1} \leq 1$, i.e., only a single record may be replaced, the following inequality holds:
\begin{equation}
{\rm Pr}\left[\mathcal{M}(d) \in \mathcal{S}\right] \leq e^{\varepsilon} {\rm Pr}\left[\mathcal{M}(d') \in \mathcal{S}\right] + \delta.
\end{equation}
$\varepsilon$ is known as the privacy budget and it bounds the worst-case privacy leakage from interactions with the data through $\mathcal{M}$. $\delta$ is the probability of failure such that the $\varepsilon$ bound holds with probability at least $1-\delta$. Thus, we care about scenarios where $\delta \ll 1/N$ where $N=|d|$ is the number of data points.

The Gaussian mechanism is perhaps the most common query in DP machine learning. This mechanism outputs Gaussian noise to the output of a (likely deterministic and) bounded function. It is defined as follows: $\mathcal{M}_{Gaussian}\triangleq g(x) + N(0, z * C)$, where $\forall x, \lVert g(x)\rVert_2 \leq  C$. Here, $C$ represents an $\ell_2$ bound on the output of $g$ which defines the ``sensitivity'' of the mechanism (in DP terms), and $z$ is known as the noise scale or noise multiplier.
\subsection{Learning with Differential Privacy}
Perhaps the most widely studied setup in both machine learning and privacy-preserving machine learning is the supervised learning scenario. Here, we have a model $f_\theta$ which is a function or `hypothesis' parameterized by $\theta$. Given a training dataset $(x^1, y^1), ..., (x^n, y^n)$ where $x^i \in \mathcal{X}$ are the inputs and $y^i \in \mathcal{Y}$ are the desired outputs (i.e., $\forall i$ we desire that $f_\theta(x^i)=y^i$), a model is learned by minimizing the empirical risk on this training data. The risk associated with a given function $f_\theta$ is characterized by a chosen loss function $\mathcal{L}$, and is defined as $R(f_\theta)=\mathrm{E}\left[\mathcal{L}\left(f_\theta\left(x\right), y\right)\right]$. In supervised machine learning, $\mathcal{L}$ is commonly the cross-entropy loss.

\paragraph{Differentially Private Empirical Risk Minimization} Chaudhuri et al.~\cite{chaudhuri2011differentially} initiated this exploration; later, Bassily et al.~\cite{bassily2014private}  showed that the excess risk must grow as $\sqrt{d}/\varepsilon$. This provides intuition that increasing the number of parameters $d$ may not always be beneficial in DP ERM and hence DP learning---though increasing $p$ enables training of larger models with more representational power, which has been shown to improve model performance in non-private settings~\cite{brown2020language}, in private settings this also increases the excess risk which may counteract this benefit. This is confirmed in practice~\cite{shen2021towards,tramer2020differentially,yu2021not,kurakin2022toward}. For example, Papernot et al. observe exactly this phenomenon when increasing the number of filters in an image classification model~\cite{papernot2019making}: though non-private performance continues to increase, the private performance decreases after some maximum.

\paragraph{Differentially Private SGD}
In machine learning, DP-SGD is the most common optimization algorithm~\cite{abadi2016deep}. 
In DP-SGD, we view the query into the dataset as the computation of the model gradients, i.e., $g$ is defined as $g = \nabla_\theta \mathcal{L}(f(X), Y)$ where $X \in \mathcal{X}, Y \in \mathcal{Y}$ are some corresponding subsets of the data, often known as minibatches. Then, to bound the sensitivity of the this query, the per-example (i.e., $\forall x \in X$) gradients are scaled to some maximum $\ell_2$ clipping norm $C$ and Gaussian noise of standard deviation $\sigma \cdot C$ is added to the sum of the per-example gradients, i.e., the minibatch gradient. $\sigma$ is calibrated to guarantee $(\varepsilon,\delta)$-DP after composing across minibatches, accounting for any privacy amplification~\cite{feldman2022hiding,PAS,bassily2014private,zhu2019poission,WangBK19}, and composing across multiple passes through the dataset, i.e., after multiple epochs of training.

%% file: method.tex
\section{Privately Fine-tuning Large Pretrained Models}

Existing work has ported fine-tuning strategies discussed in the non-private literature to privately fine-tune large pretrained models. Prior to these efforts, the accuracy gap between learning privately and non-privately was prohibitive and prevented the adoption of differentially private learning~\cite{papernot2021tempered,tramer2020differentially}. Generally speaking, the approach taken by two representative recent papers due to De et al.~\cite{de2022unlocking} and Mehta et al.~\cite{mehta2022large} involves pretraining a model on a first dataset, termed the pre-training dataset, and then fine-tuning its entire parameter vector on a second dataset, termed the downstream dataset. This is one of the earliest methods for non-private fine-tuning~\cite{erhan2010does}. Adapting this strategy to private learning, De et al. and Mehta et al. were able to scale training with DP-SGD to datasets like CIFAR10 and ImageNet. 

\paragraph{Background.} De et al.~\cite{de2022unlocking} fine-tune a Wide-ResNet~\cite{zagoruyko2016wide} and NFNet-F3\cite{brock2021high}. For the CIFAR10 and CIFAR100 downstream tasks, they find that fine-tuning all layers performs best. For the ImageNet downstream task, they instead find that fine-tuning the last layer performs better. Outside of these two approaches to fine-tuning, little attention was paid to the choice of layers for fine-tuning. Instead, the authors focus on discussing a number of improvements they make to their implementation of DP-SGD to increase accuracy given a fixed privacy guarantee. This includes using group normalization, large minibatch sizes (e.g., 4096 or 16,384 points per minibatch), applying weight standardization, and using exponential moving averages for parameter averaging. Furthermore, they introduce the concept of augmentations to DP-SGD: they find that computing the average gradient over augmented examples, before the gradient is clipped so as to not impact the privacy guarantee, yields increased performance.

Concurrently, Mehta et al.~\cite{mehta2022large} fine-tune vision transformers (ViTs)~\cite{dosovitskiy2020image}, in particular ViT-H/14. For the ImageNet downstream task, they also only consider two settings: fine-tuning all layers or the last layer only. When using DP-SGD, they find that fine-tuning the last layer only performs best---confirming the findings of De et al~\cite{de2022unlocking}. They also find that large minibatches perform best (e.g., 128k to 1M points per minibatch). In fact, their best result is obtained for batch gradient descent: i.e., taking only a single step of gradient descent on the entire dataset used for fine-tuning. 

\paragraph{Our contribution.} \textit{In this paper, we ask whether these results can be further improved by carefully tailoring the fine-tuning procedure to the specifics of learning with differential privacy.} We answer the question in the affirmative.

Specifically, we explore the impact on performance that the different model parameters being fine-tuned have. 
Take the example of non-private fine-tuning. While there is no principled approach to reason about this to the best of our knowledge~\cite{he2022parameter}, established practices either fine-tune the entire pretrained model or they focus on fine-tuning the last layer(s) of the pretrained model. This is based on the intuition of general and task-specific features, where  features learned by layers that are lower in the architecture are typically more general~\cite{yosinski2014transferable}. Following this, the most common approaches were to fine-tune either the last~\cite{donahue2014decaf} or the last few~\cite{long2015learning} layers---that learn features that are more task-specific. At times, the last layer will be learned from a random initialization or a new linear layer is appended to the model~\cite{radford2021learning}. Given the lack of existing research on this aspect, we argue that the question of which parameters need to be fine-tuned to achieve optimal performance merits its own investigation in the context of private fine-tuning. 

\paragraph{Our approach.} Upon experimenting with different configurations of fine-tuning, \textbf{we propose that the first and last layer of the pretrained model be fine-tuned when training with DP-SGD.} We observe in Section~\ref{sec:results} that this yields significant improvements in accuracy compared to existing state-of-the-art results that fine-tune the entire pretrained model. Because this choice reduces the norm of the noise added to obtain DP, we hypothesize that it enables us to alleviate some of the tensions between privacy and scale, while fine-tuning the layers that are more important for high performance on the task: both the later task-specific layers along with the earlier more general features.\footnote{We are further investigating this hypothesis and will include an extended discussion in a full version of this manuscript.}

%% file: experiments.tex
\section{Main Experimental Result}
\label{sec:results}

To evaluate our approach, we compare our approach to the baselines established by the current state-of-the-art approach from De et al.~\cite{de2022unlocking}. We use the same experimental setup.

\paragraph{Datasets and Architectures.} We include an evaluation on two common computer vision benchmarks: CIFAR-10 and CIFAR-100~\cite{krizhevsky2009learning}. %
On these datasets, we fine-tune a 28-10 Wide-ResNet which was initially pretrained on Imagenet. %

\paragraph{Comparing with De et al.~\cite{de2022unlocking}} 
We follow the many optimizations introduced by De et al.~\cite{de2022unlocking} to achieve near non-private performance, even under the reasonable privacy budget of $2,10^{-5}$-DP. We then switch their whole or last-layer fine-tuning step with our first-last-layer fine-tuning step, with no other modifications.

\paragraph{Main result.} Our main result is that first-last-layers fine-tuning consistently improves on all prior state-of-the-art results that used either whole-model or last-layer fine-tuning. For instance, we achieve 77.9\% accuracy for $(\varepsilon, \delta)=(2, 10^{-5})$ on CIFAR-100 compared to $74.6\%$ when De et al. fine-tuned the entire model.  

\begin{table}[H]
\caption{\textbf{First-last-layers fine-tuning consistently improves on all prior results using either whole-model or last-layer fine-tuning.} We observe that when the model has already plateaued near the non-private baseline, there are no additional gains from using first-last-layer fine-tuning; otherwise, we observe gains up to $3.4$ percentage points on CIFAR-100 for $\varepsilon=1$. Results reported are either taken from~\cite{de2022unlocking} or recreated using their exact setup.}
\label{tab:scale_regime_and_optimizer}
    \centering
    \begin{tabular}{cccccccc}
    \toprule
        \multirow{2}{*}{Dataset} & \multirow{2}{*}{Method} & \multirow{2}{*}{~} & \multicolumn{4}{c}{Top-1 DP Fine-Tuning Accuracy (\%)}\\ %
        \cmidrule{4-7}
        & & & $\epsilon=1$ & $\epsilon=2$ & $\epsilon=4$ & $\epsilon=8$\\
        \toprule
        CIFAR10 & \multirow{3}{*}{\shortstack{Whole-Model\\Last-Layer\\\shortstack{First-Last-Layers (Ours)}}} & & 94.7 &
        95.4 & \textbf{96.1} & \textbf{96.7} \\
         & & & 93.1 & 93.6
         & 94.0 & 94.2 \\
        & & & \textbf{95.0} &
        \textbf{95.6} & \textbf{96.1} & 96.4\\\hline
        CIFAR-100 & \multirow{3}{*}{\shortstack{Whole-Model\\Last-Layer\\\shortstack{First-Last-Layers (Ours)}}} & & 70.3 & 74.7 & 79.2 & 81.8\\
        & & & 70.3 & 73.9 & 76.1 & 77.6 \\
        & & & \textbf{73.7} & \textbf{77.9} & \textbf{81.0} & \textbf{82.1}\\
        \bottomrule
    \end{tabular}
    \vskip -0.1in
\end{table}

%% file: discussion.tex
\section{Conclusion}

The results we presented in this short note call for an additional hyperparameter search to be completed when fine-tuning with differential privacy. 
However, because of the potential for privacy leakage from hyperparameters~\cite{papernot2022hyperparameter}, many of these fine-tuning strategies introducing additional hyperparameters cannot be directly applied to the private setting. Our work explores the effectiveness of these strategies in the private setting, especially when considering the tensions between privacy and scale.